\pdfoutput=1

\documentclass[11pt]{article}

\usepackage[preprint]{acl}

\usepackage{booktabs}       
\usepackage{times}
\usepackage{latexsym}
\usepackage{amsmath}   
\usepackage{amssymb}   
\usepackage{multirow}
\usepackage{makecell}
\usepackage{enumitem}
\usepackage{hyperref}
\usepackage[linesnumbered,ruled,vlined]{algorithm2e}
\usepackage{float}

\usepackage[T1]{fontenc}

\usepackage[utf8]{inputenc}

\usepackage{microtype}
\usepackage{enumitem}

\usepackage{inconsolata}

\usepackage{graphicx}

\title{Bitnet.cpp: Efficient Edge Inference for Ternary LLMs}

\author{
 \textbf{Jinheng Wang\textsuperscript{1,4}},
 \textbf{Hansong Zhou\textsuperscript{1,4}},
 \textbf{Ting Song\textsuperscript{4}},
 \textbf{Shijie Cao\textsuperscript{4}},
 \textbf{Yan Xia\textsuperscript{4}},
 \textbf{Ting Cao\textsuperscript{4}},
 \\
 \textbf{Jianyu Wei\textsuperscript{2,4}},
 \textbf{Shuming Ma\textsuperscript{4}},
 \textbf{Hongyu Wang\textsuperscript{3,4}},
 \textbf{Furu Wei\textsuperscript{4}},
\\
 \textsuperscript{1}Peking University,
 \textsuperscript{2}University of Science and Technology of China,
\\
 \textsuperscript{3}University of Chinese Academy of Sciences,
 \textsuperscript{4}Microsoft Research
\\
}

\begin{document}
\maketitle
\begin{abstract}
The advent of 1-bit large language models (LLMs), led by BitNet b1.58, has spurred interest in ternary LLMs. Despite this, research and practical applications focusing on efficient edge inference for ternary LLMs remain scarce. To bridge this gap, we introduce Bitnet.cpp, an inference system optimized for BitNet b1.58 and ternary LLMs. Given that \textbf{mixed-precision matrix multiplication (mpGEMM)} constitutes the bulk of inference time in ternary LLMs, Bitnet.cpp incorporates a novel mpGEMM library to facilitate sub-2-bits-per-weight, efficient and lossless inference. The library features two core solutions: \textbf{Ternary Lookup Table (TL)}, which addresses spatial inefficiencies of previous bit-wise methods, and \textbf{Int2 with a Scale (I2\_S)}, which ensures lossless edge inference, both enabling high-speed inference. Our experiments show that Bitnet.cpp achieves up to a 6.25x increase in speed over full-precision baselines and up to 2.32x over low-bit baselines, setting new benchmarks in the field. Additionally, we expand TL to element-wise lookup table (ELUT) for low-bit LLMs in the appendix, presenting both theoretical and empirical evidence of its considerable potential. Bitnet.cpp is publicly available at \url{https://github.com/microsoft/BitNet/tree/paper}, offering a sophisticated solution for the efficient and practical deployment of edge LLMs.
\end{abstract}

\section{Introduction}
In recent years, large language models have garnered widespread attention due to their exceptional performance across a variety of tasks. However, the growing demand for efficient deployment on edge devices, particularly driven by data privacy concerns, poses challenges due to the limited computational power and bandwidth of these devices.

\begin{figure}[htbp]
    \centering
    \includegraphics[width=\columnwidth]{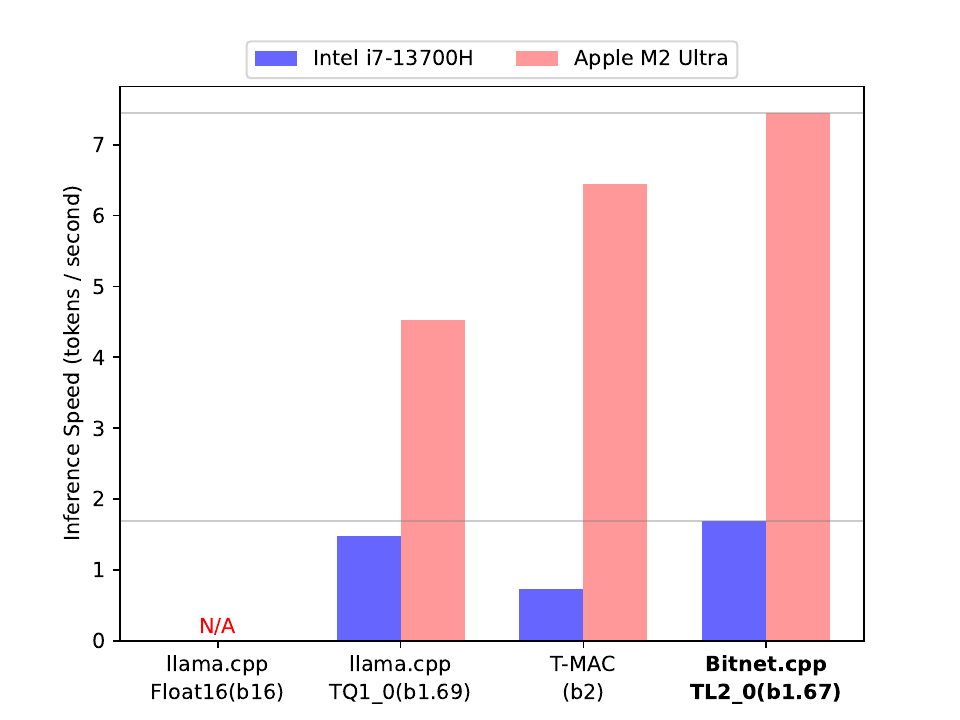}
    \caption{A comparison of end-to-end inference speeds on a 100B ternary LLM. $(bx)$ represents the bits per weight, where $x$ denotes specific value. "N/A" indicates that the tested CPU cannot host the specified model size with the given kernel.}
    \label{fig:title}
\end{figure}


\begin{figure*}[htb]
    \centering
    \includegraphics[width=\linewidth, trim=10 210 10 110, clip]{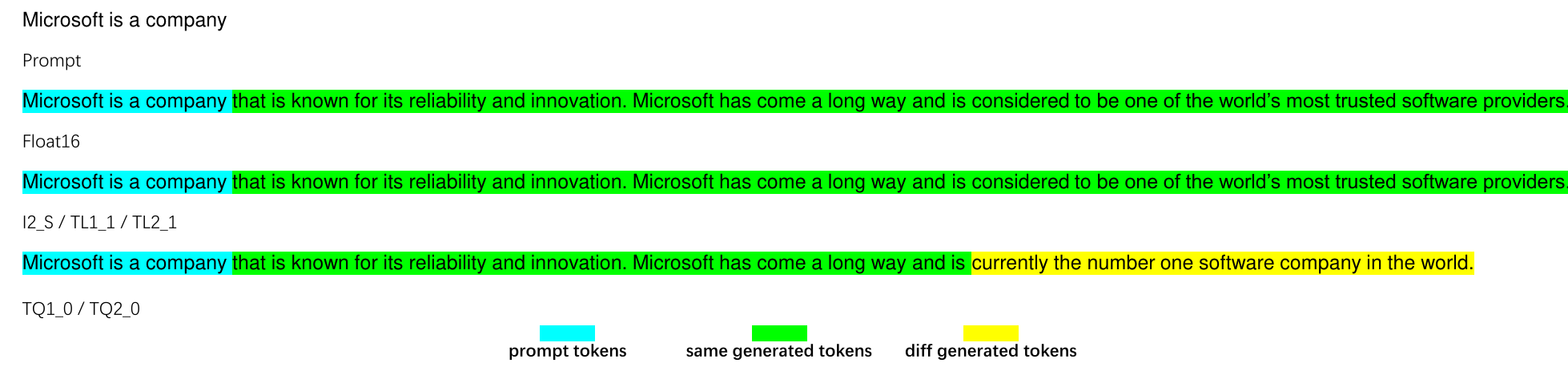}
    \caption{\label{fig:lossless}An example to demonstrate lossless inference for BitNet b1.58 with Bitnet.cpp.}
\end{figure*}

To address these challenges, model compression techniques are frequently employed. Notable examples benefiting from such techniques include Gemini-Nano \cite{gemini-nano} and Phi3-mini \cite{phi3}, both designed for mobile and personal devices. Furthermore, recent advancements by BitNet b1.58 \cite{wang2023bitnet, ma2024era} represent a significant development in edge LLM inference, introducing 1-bit LLMs by quantizing all weights to ternary values therefore reducing the bits per weight (bpw) to 1.58, while preserving accuracy comparable to full-precision LLMs. Subsequent releases of ternary LLMs, including TriLM \cite{trillm}, Llama3-8B-1.58 \cite{llama3-1.58b}, and BitNet a4.8 \cite{wang2024bitneta484bitactivations}, have demonstrated the effectiveness and applicability of ternary architectures, thereby extending the boundaries of the 1-bit era. Despite the burgeoning interest in ternary LLMs, the conversion of their theoretical benefits into practical advantages during inference is still understudied.

To fill this gap, we aim to enhance the performance of ternary LLMs edge inference by optimizing mpGEMM (e.g., 8-bit activation and 1.58-bit weight). However, the non-integer bpw characteristic of ternary weights conflicts with the rules for computer memory access alignment, thus posing challenges in designing a sub-2-bit-per-weight, efficient edge mpGEMM for ternary LLMs. Currently, TQ1\_0 in llama.cpp\cite{llamacpp} utilizes 1.69 bits to store ternary weights, but it is slower compared to TQ2\_0 and T-MAC\cite{wei2024t}, which use 2 bits to maintain alignment. Moreover, prior implementations of mpGEMM have not achieved lossless inference for BitNet b1.58, as they fail to fully align with BitNet b1.58 training schemes during inference.

To address these issues, we develop Bitnet.cpp, which incorporates a novel mpGEMM library. Our key idea is to avoid intricate bit-level manipulations by directly operating the weight elements when designing mpGEMM, while strictly aligning with BitNet b1.58 training schemes. Based on our ideas, the library not only resolves spatial inefficiencies, but also surpasses existing solutions in terms of performance (Figure \ref{fig:title}), achieving lossless inference for BitNet b1.58 (Figure \ref{fig:lossless}). To this end, our work makes several contributions:

\begin{itemize}[left=0pt]
\item
First, we conduct a comprehensive survey of current cutting-edge mpGEMM methods and identify their limitations when applied to ternary LLMs. (Section \ref{sec:2})
\item 
To overcome these limitations, we design and implement a ternary mpGEMM library incorporating our innovative kernels, \textbf{TL} and \textbf{I2\_S}, which we integrate into Bitnet.cpp. This library facilitates fast and lossless inference through element-wise design and precise alignment with training schemes. (Section \ref{sec:3})
\item
We evaluate Bitnet.cpp on multiple edge devices and demonstrate that it achieves a up to 6.25x speedup compared to state-of-the-art baselines, while realizing lossless inference for BitNet b1.58. (Section \ref{sec:4})
\item
Finally, in the appendix, we extend TL beyond ternary LLMs, to element-wise lookup table (ELUT) for low-bit LLMs. We perform both theoretical (Appendix \ref{sec:insight}) and practical (Appendix \ref{sec:analy}) analyses of ELUT, demonstrating its high efficiency and untapped potential. (Appendix \ref{sec:poten}).

\end{itemize}

\section{\label{sec:2}Ternary LLM \& mpGEMM on Edge}
\begin{figure}[htbp]
    \centering
    \includegraphics[width=\columnwidth, trim=180 50 100 80, clip]{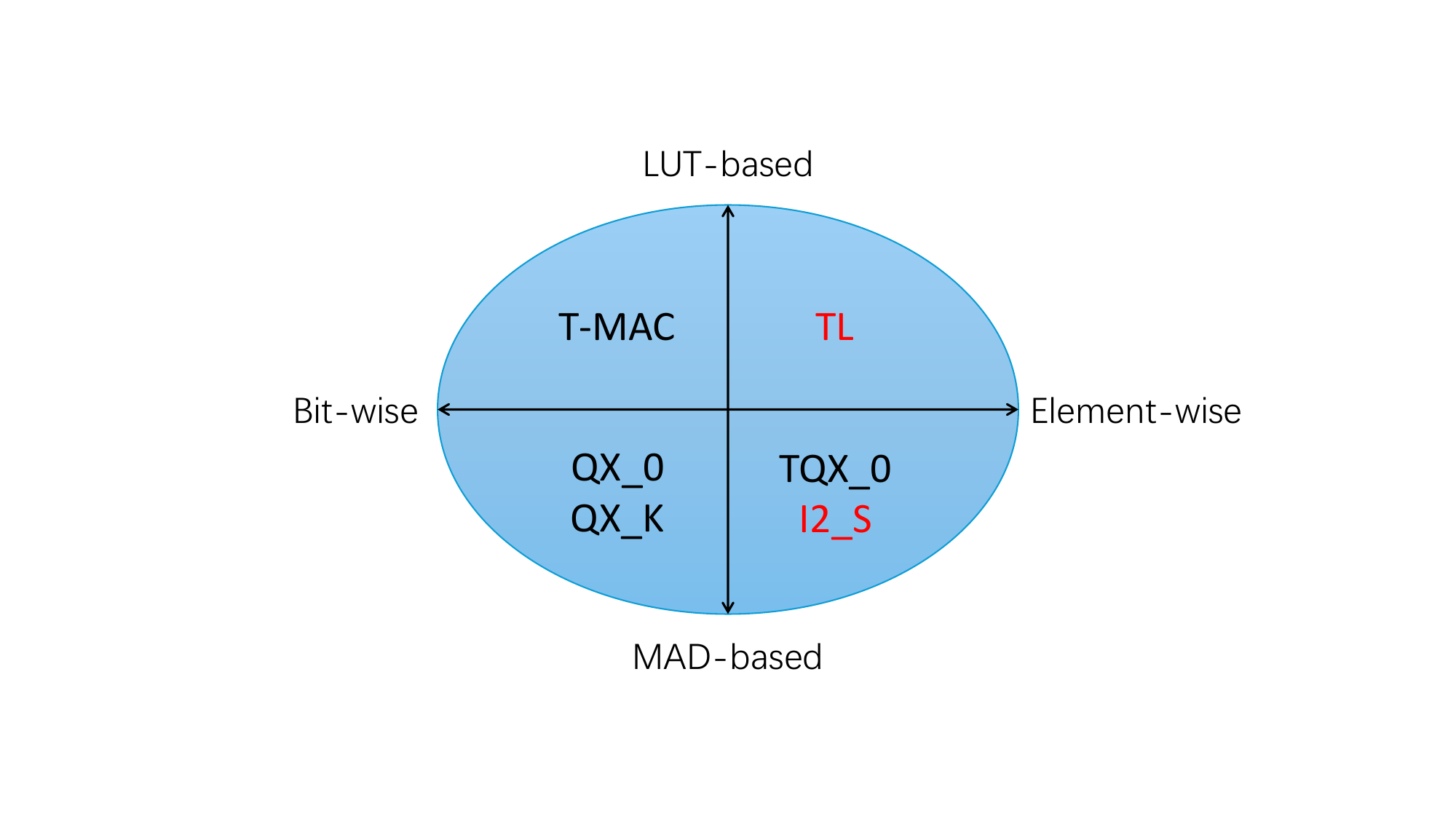}
    \caption{\label{fig:classfication}A taxonomy of mpGEMM solutions for ternary LLMs on edge devices. TL and I2\_S are integrated in Bitnet.cpp, while QX and TQX are integrated in llama.cpp.}
\end{figure}

In this section, we present a detailed examination of the characteristics of ternary LLMs and introduce a systematic taxonomy of current edge mpGEMM methods, as illustrated in Figure \ref{fig:classfication}. We aim to delineate the limitations of existing mpGEMM approaches in handling ternary LLMs, informed by our comprehensive survey, with the objective of guiding future optimizations.

\subsection{Ternary LLM: Features}

\noindent \textbf{Ternary Weights}
A distinctive characteristic of ternary LLMs is that the weights in the transformer layers are ternary, allowing only three possible values: \{-1, 0, 1\}. Consequently, the information content of these weights is approximately 1.58 bits per weight, as calculated by \( \log(3) / \log(2) \). This substantial compression not only markedly reduces the model size, but also enables opportunities for further optimization with existing mpGEMM methods, such as those employed in llama.cpp and T-MAC.

\noindent \textbf{Lossless Inference for BitNet b1.58}
BitNet b1.58 performs ternary quantization on weights and int8 per-tensor quantization on activations during training. Based on this, if the training constraints are preserved during inference, lossless inference can be achieved for BitNet b1.58, as shown in Figure \ref{fig:lossless}.

\subsection{mpGEMM on Edge: Definitions}
\noindent \textbf{MAD-based and LUT-based}
We classify edge mpGEMM methods into two computational strategies: \textbf{multiply-then-add (MAD)-based} and \textbf{lookup table (LUT)-based}. The MAD-based strategy performs dot product calculations, while the LUT-based strategy employs lookup tables to store precomputed values, thereby enabling rapid accumulation via table lookups.

\noindent \textbf{Bit-wise and Element-wise}
Edge mpGEMM methods are additionally classified based on the fundamental unit of computation into \textbf{Bit-wise} and \textbf{Element-wise} categories. Bit-wise methods process data at the bit level, focusing solely on bit operation without considering the attributes of weight elements, precluding non-integer bits per weight. In contrast, element-wise methods perform computations at the element level, taking into account the distinct properties of each weight element, which enables non-integer bits per weight.

\subsection{mpGEMM on Edge: Taxonomy (Figure \ref{fig:classfication})}
\noindent \textbf{Bit-wise LUT-based (Up left)}
Recent research by T-MAC has shown that bit-wise LUT-based methods significantly outperform MAD-based approaches in edge inference, particularly emphasizing their efficiency for low-bit LLMs. However, when applied to ternary LLMs, these bit-wise LUT-based methods exhibit spatial inefficiencies, leading to a substantial performance decline in environments with limited bandwidth.

\noindent \textbf{Bit-wise MAD-based (Down left)}
As a foundational framework for LLM edge inference, llama.cpp has pioneered several low-bit edge mpGEMM methods, predominantly bit-wise MAD-based, including the QX\_0 and QX\_K series. For instance, Q2\_K utilizes the K-quants method to quantize weights to 2 bits, thereby ensuring the universality and correctness of the quantization. However, the application of Q2\_K to ternary weights introduces complications: in addition to wasted space, maintaining accuracy with K-quants necessitates a multi-step dequantization process prior to performing the dot product, consequently increasing the overall latency.

\noindent \textbf{\label{sec:losslessdef}Element-wise MAD-based (Down right)}
In fact, llama.cpp introduces two element-wise MAD-based methods for ternary LLMs: TQ1\_0 and TQ2\_0, with bits per weight of 1.69 and 2.06, respectively. These methods leverage the ternary nature of the weights to avoid the multi-step dequantization required by K-quants, thereby significantly boosting performance. Despite these advancements, the lack of support for tensor-wide quantization means llama.cpp relies on per-block quantization with a static block length of 256 for activations (e.g., Q8\_K). To accommodate this limitation, TQX\_0 also utilizes the block quantization scheme. However, this approach is inconsistent with the computational methods used during BitNet b1.58 training, thus hindering TQX\_0 from achieving lossless inference.

\begin{figure*}[t]
  \includegraphics[width=\linewidth, trim=0 245 0 45, clip]{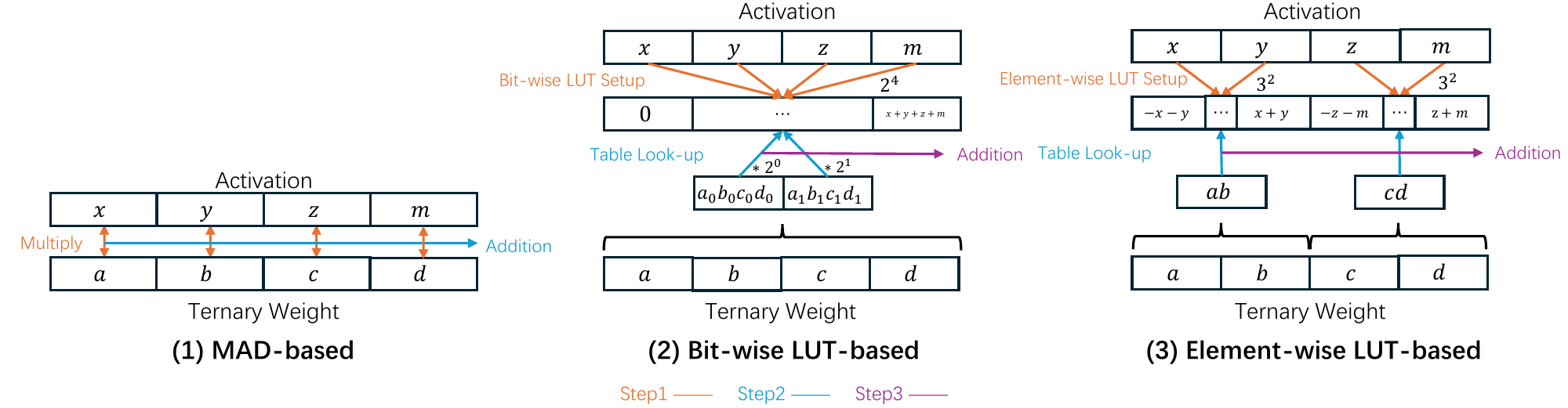}
  \caption {\label{fig:diff}A simple example to explain the differences between various methods for completing mpGEMM when $K=4$: (1) represents the MAD-based solution, where the result is obtained via the dot product; (2) represents the bit-wise LUT-based solution, where the weights are split into different bit indices, and the result is obtained by performing a lookup in the LUT, followed by bit-shifting and accumulation; (3) represents the element-wise LUT-based solution, where all possible values of the weights are enumerated to obtain the index, and the result is obtained by performing a lookup in the LUT, followed by accumulation. $A_x$ refers to the $x_{th}$ bit in weight $A$. In (2), $g = 4$ and $b = 2$; whereas in (3) $g = 2$ and $C = 3$.}
\end{figure*}

\section{\label{sec:3}Methodology}
\begin{table}[ht]
\centering
\begin{tabular}{|c|c|c|c|c|}
\hline
Kernel & type & bpw & Lossless\\
\hline
TL1\_0 & LUT-based & 2 & $\times$ \\
\hline
TL1\_1 & LUT-based & 2 & $\checkmark$ \\
\hline
TL2\_0 & LUT-based & 1.67 & $\times$ \\
\hline
TL2\_1 & LUT-based & 1.67 & $\checkmark$ \\
\hline
I2\_S & MAD-based & 2 & $\checkmark$ \\
\hline
\end{tabular}
\caption{\label{tab:tl}Bitnet.cpp ternary mpGEMM library.}
\end{table}

This section addresses the limitations of existing edge mpGEMM methods, as previously discussed, through the design and implementation of a novel ternary mpGEMM library, summarized in Table \ref{tab:tl}. We aim to showcase our pioneering techniques for efficient edge inference of ternary LLMs, focusing on two key dimensions: fast and lossless.



\subsection{Fast Edge Inference}
For MAD-based methods, llama.cpp has implemented TQ1\_0 and TQ2\_0, which facilitate rapid ternary LLM edge inference. However, the current bit-wise approach for LUT-based methods does not fully exploit the potential of ternary LLMs for fast edge inference. Consequently, we have developed the \textbf{element-wise LUT-based (ELUT) mpGEMM}, which not only reduces bpw but also addresses the spatial inefficiencies inherent in bit-wise methods through \textbf{element-wise mirror consolidation}. To effectively implement ELUT in ternary LLMs, noted as \textbf{TL}, we mitigate issues such as misaligned memory access through \textbf{signed-unsigned weight splitting}, overcome hardware instruction support deficiencies with \textbf{1bit sign operation}, and resolve misaligned block computations via \textbf{block-fitting weight splitting}. This subsection will elaborate on our design and implementation strategies. For an in-depth analysis of the reasons behind ELUT's acceleration and its broader implications beyond ternary LLMs, please refer to Appendix \ref{sec:insight}.
\subsubsection{Design: TL}
\textbf{Element-wise LUT-based mpGEMM} 
The bit-wise LUT-based mpGEMM, designed for generality, uses 2-bit storage for ternary weights, leading to space inefficiency, thus negatively affecting speed. To overcome these limitations, we introduce an element-wise LUT-based mpGEMM approach. In the following, we delineate the key distinctions among MAD-based, bit-wise LUT-based, and element-wise LUT-based mpGEMM methods.
\begin{equation}
  \label{eq1:example}
  R = \sum_{i \gets 1}^{K}A_iW_i
\end{equation}
\begin{equation}
  \label{eq2:example}
  R = \sum_{i \gets 1}^{b}\sum_{j \gets 1}^{K/g} \text{Look-up}(bLUT_j, W_{ij})
\end{equation}
\begin{equation}
  \label{eq3:example}
  R = \sum_{i \gets 1}^{K/g} \text{Look-up}(eLUT_i, W_i)
\end{equation}
\begin{equation}
  \label{eq4:weight}
  W \in \mathbb{Z}, \ |W| = C
\end{equation}

Consider a simple GEMM computation involving two input matrices: \( A \) (1, K) and \( B \) (K, 1). As shown in Equation~\ref{eq1:example}, MAD-based mpGEMM computes the result using the dot product. In LUT-based mpGEMM, the conventional approach employs a bit-wise representation of the LUT, as shown in Equation~\ref{eq2:example}, where \( b \) denotes the bit-width of the weight (2 for ternary weights, as \( 3 < 2^2 \)), and \( g \) represents the group size. The bit-wise LUT (\( bLUT \)) has a size of \( b^g \). By relaxing the bit-width restriction and adopting an element-wise representation of the LUT, as shown in Equation~\ref{eq3:example}, a finer-grained expression is obtained. In this case, the element-wise LUT (\( eLUT \)) has a size of \( C^g \), where \( C \) denotes the cardinality of the weight set (\( 3 \) for ternary weights). Figure~\ref{fig:diff} illustrates a simple example highlighting these differences.

\begin{figure*}[htb]
    \centering
    \includegraphics[width=\linewidth, trim=30 10 100 10, clip]{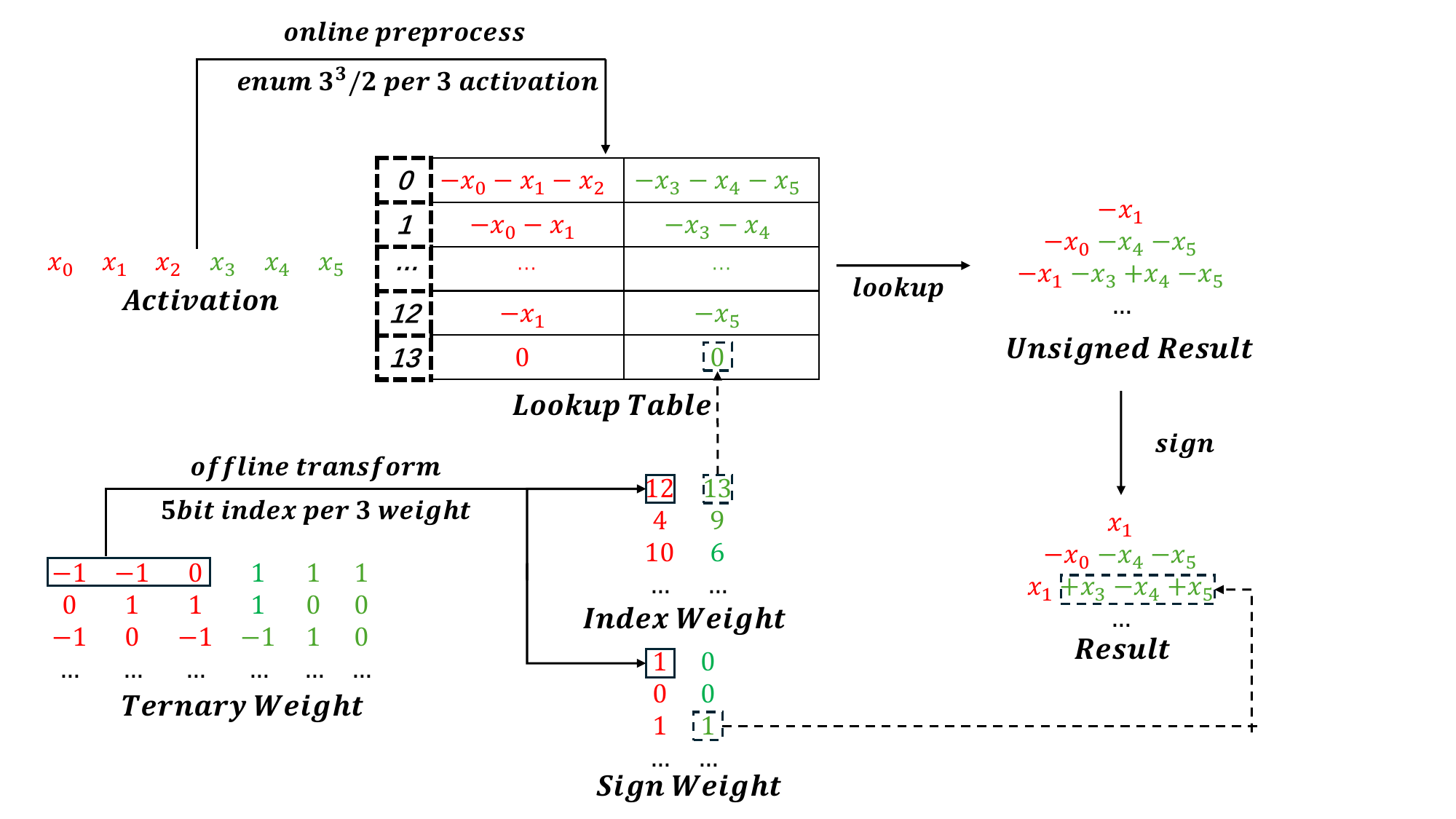}
    \caption{\label{fig:tl2-design}The TL2 design uses signed-unsigned weight splitting. First, a 4-bit index weight is used to look up the table and obtain the unsigned result. Then, the corresponding 1-bit sign weight is applied to perform the sign operation on the unsigned result, yielding the final output.
}
\end{figure*}


\noindent \textbf{Element-wise Mirror Consolidation}
\cite{wei2024t} introduced the concept of mirror consolidation, positing that during LUT enumeration, half of the values for $b^g$ are inversely related to the other half, effectively halving the LUT size. Extending this concept to $C^g$ results in what we term element-wise mirror consolidation. For the element-wise LUT-based solution, due to the 128-bit SIMD register instruction length (e.g., AVX2 vpshufb), \( C^g \) is constrained to a maximum of 16 ($16 \times int8 = 128$). Without element-wise mirror consolidation, the maximum value of \( g \) for ternary LLMs remains at 2, maintaining the same bpw as the bit-wise LUT-based method (4 bits for 2 weights, \( 3^2 < 2^4 \)). However, employing element-wise mirror consolidation increases the maximum \( g \) to 3, thus compressing bpw to 1.67 (5 bits for 3 weights, \( \frac{3^3}{2} < 2^4 \)). Consequently, we have developed two practical designs for TL. We refer to the design with $g = 2$ as TL1 and the design with $g=3$, which incorporates element-wise mirror consolidation, as TL2. Algorithm~\ref{alg:TL1} details the design of TL1, while Algorithm~\ref{alg:TL2} outlines that of TL2.

\subsubsection{Implementation: TL}
\textbf{Signed-Unsigned Weight Splitting}
To implement element-wise mirror consolidation, we introduce signed-unsigned weight splitting, where we use a separate 1-bit sign weight to store the sign of the enumeration, and a 4-bit index weight to store the corresponding LUT index for unsigned enumeration. It is evident that using continuous 5-bit storage for 3 weights would cause severe memory access misalignment. Since LUT-based mpGEMM is inherently memory-intensive, the additional memory accesses caused by misalignment would significantly degrade performance. In contrast, signed-unsigned weight splitting allows three weights to be represented using 5 bits, adhering to the element-wise approach, while simultaneously avoiding misalignment issues in computation and memory access. Figure~\ref{fig:tl2-design} demonstrates the detailed computation flow of TL2, using signed-unsigned weight splitting.

\noindent \textbf{1bit Sign Operation}
Determining the sign of the value indexed from the LUT using only 1 bit is challenging, as values are represented in two's complement, and the design must ensure compatibility with SIMD instructions. 
\begin{equation}
\begin{aligned}
    x = \text{sign} &\oplus (\text{sign} + x) \\
    x \in  \text{int8}&,\  \text{sign} \in \{0,1\} \\
\end{aligned}
\label{eqn:sign}
\end{equation}
After evaluating multiple methods, we selected the approach presented in Equation~\ref{eqn:sign} to address the issue. This sequence of operations, which includes the XOR and ADD operations, enables the sign to be determined by a single bit and is fully compatible with both the AVX2 and NEON instructions. When the bit of sign is 0, the result remains unchanged; otherwise, the result is converted to its negative value.

\begin{figure*}[htb]
    \centering
    \includegraphics[width=\linewidth, trim=0 120 0 70, clip]{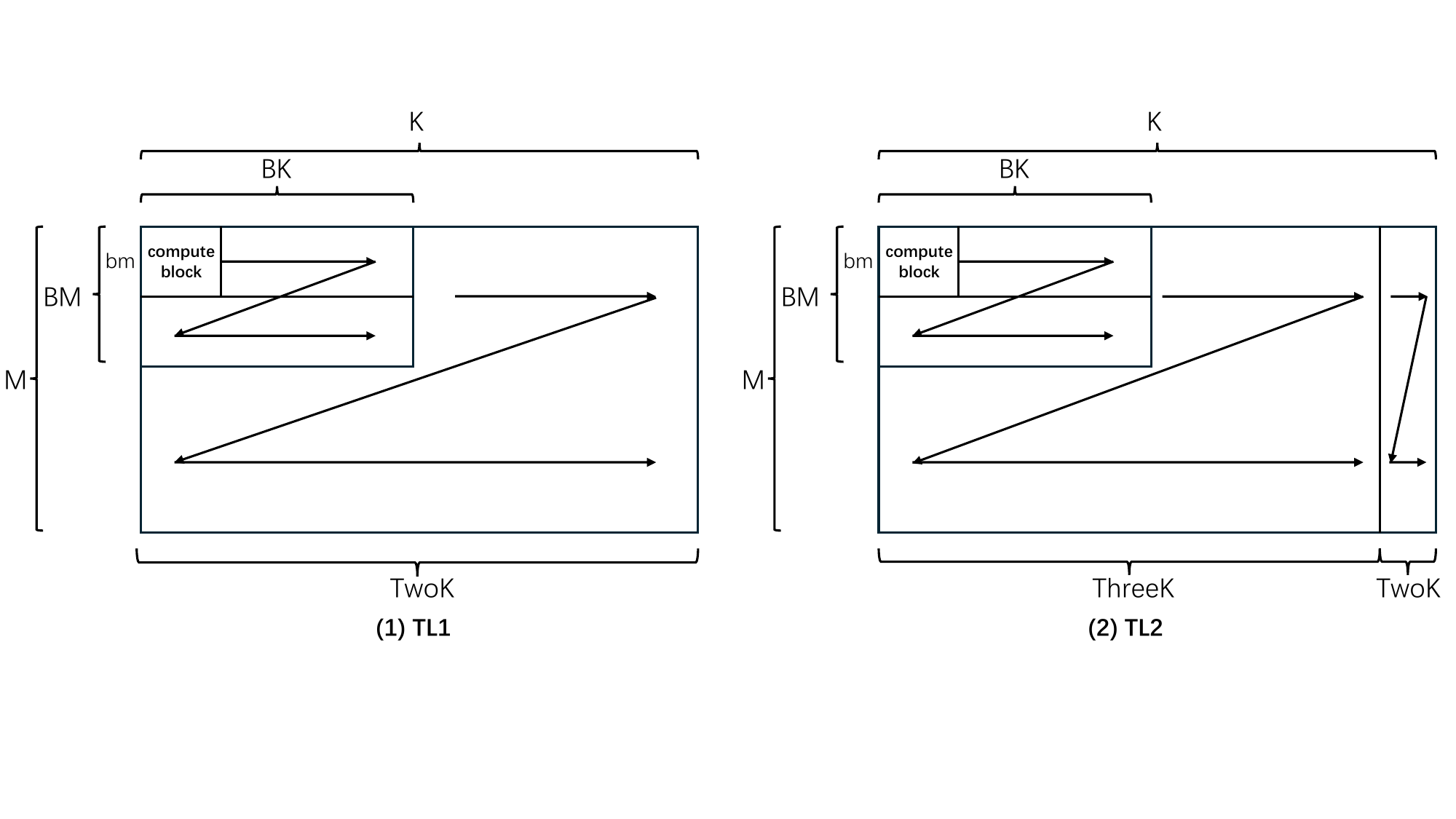}
    \caption{\label{fig:padding}The computation sequences for TL1 and TL2. The left side represents TL1, and the right side represents TL2. Arrows indicate the order of computation, with the smallest computational unit being the compute block. $M$ and $K$ refer to the dimensions of the weights. \( bm \times by \) refers to the number of weights involved in the minimal compute block. \( bm \) can be selected from 16 or 32. In TL1, \( by \) is \( \frac{256}{bm} \), and in TL2, \( by \) is \( \frac{192}{bm} \).
}
\end{figure*}

\noindent \textbf{Block-fitting Weight Splitting}
The TL series employs an LUT-centric data layout for mpGEMM to address inefficiencies in memory storage and access, as introduced by T-MAC. When adopting this layout, it is crucial to ensure that the minimal compute blocks align precisely with the weight matrix. As illustrated on the left side of Figure~\ref{fig:padding}, for TL1, the block length \( BK \) must be divisible by the matrix dimension \( K \). This condition is easily met in TL1, as \( g = 2 \), meaning \( K \) only needs to be a multiple of 2. However, the situation differs for TL2. Most LLM weight shapes do not have \( K \) as a multiple of 3 when using TL2, where \( g = 3 \). To address this, we introduce block-fitting weight splitting, which statically divides the weight into two parts to fit the blocks. After splitting, as shown on the right side of Figure~\ref{fig:padding}, one portion of the weight, with dimensions \( ThreeK = \lfloor\frac{K}{BK3}\rfloor \times BK3 \), is computed using TL2, while the remaining portion, \( TwoK = K - ThreeK \), is computed using TL1. By applying block-fitting weight splitting, we resolve the block mismatch issue without requiring additional padding, thereby preventing potential latency increases.

\begin{figure*}[ht]
    \centering
    \includegraphics[width=\linewidth]{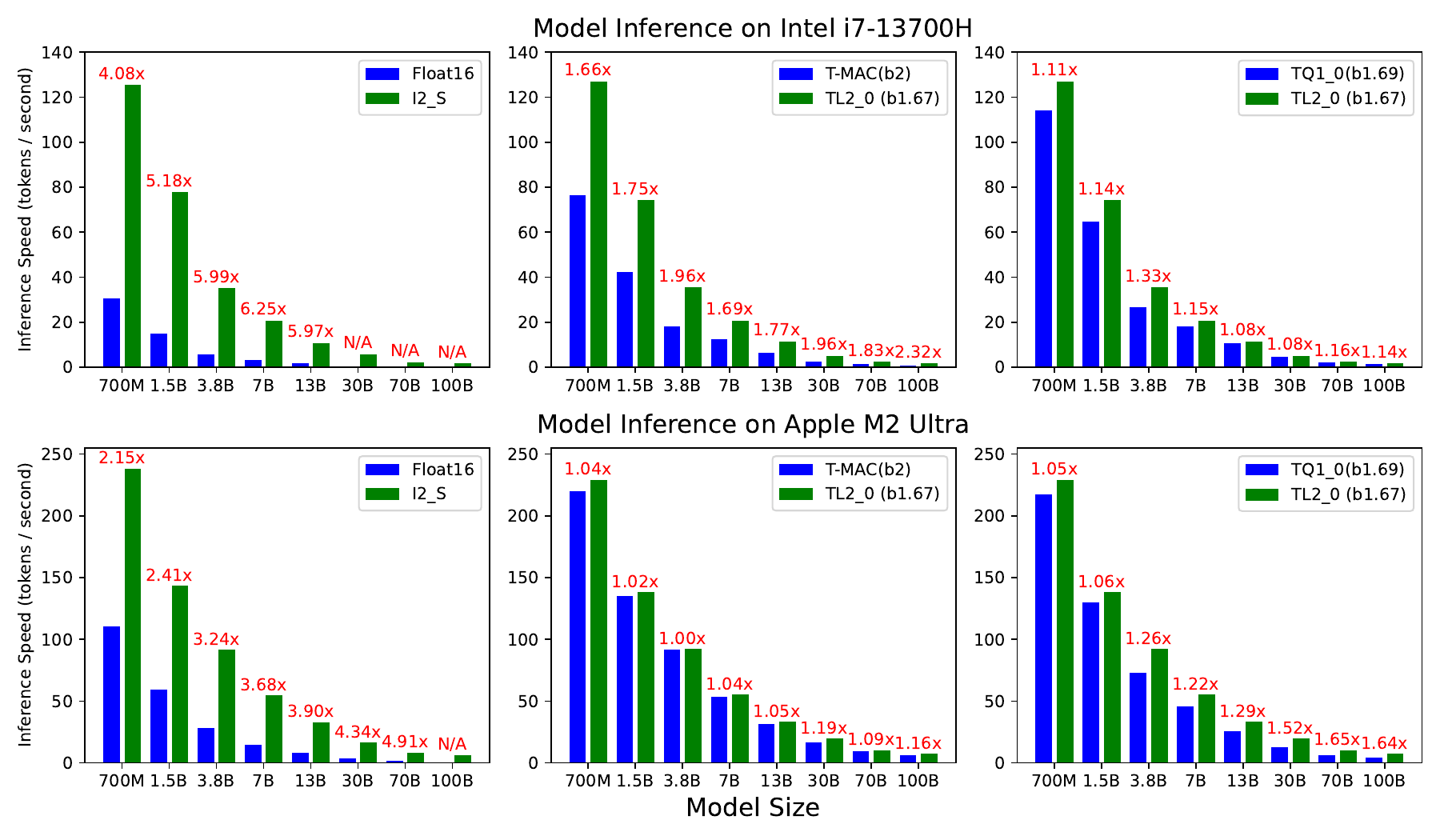}
    \caption{\label{fig:whole-compa}End-to-end performance for inference on multiple model sizes, specific details of the model size are referenced in \cite{bitnetcpp}. Here, $(bx)$ denotes the bpw value, where $x$ represents the respective bpw. Detailed performance information can be found on Table \ref{tab:detailed_com}.}
\end{figure*}

\subsection{Lossless Edge Inference}
To achieve lossless inference for BitNet b1.58, this subsection first identifies the gaps between existing methods and lossless inference. It then presents innovative approaches for achieving lossless inference using both MAD-based and LUT-based methods.

\subsubsection{Design \& Implementation: TL}
Since table lookups require SIMD instructions operating on 8-bit data, a potential conflict arises when enumerating sums that might overflow if stored in 8-bit integers. T-MAC addresses this issue by quantizing the accumulated sum to int8; however, this approach introduces additional losses, preventing lossless inference. To resolve this, we introduce the pack-and-unpack technique. First, we maintain the sums as int16 without additional quantization and split the int16 enumerated sums into two parts using the pack instruction. Then, during the indexing process, we apply the table lookup twice. Afterward, we use the unpack instruction to concatenate the two parts, ultimately obtaining the desired int16 result. Kernels that utilize typical additional quantization are TL1\_0 and TL2\_0, whereas those that use the pack-and-unpack technique are TL1\_1 and TL2\_1.

\subsubsection{Design \& Implementation: I2\_S}
Due to inconsistency with training schemes, existing element-wise MAD-based methods do not enable lossless inference for BitNet b1.58. In Bitnet.cpp, I2\_S is designed based on the element-wise approach, adhering strictly to the ternary weight and per-tensor int8 activation quantization settings of BitNet b1.58 training, thereby ensuring lossless inference. Furthermore, I2\_S performs comparably with TQ2\_0 and supports mpGEMM dimensions $K$ that are multiples of 128, while TQ2\_0 only supports multiples of 256. As a result, we have optimized the MAD-based solutions and integrated the implementation into Bitnet.cpp.

\section{\label{sec:4}Experiments}
We evaluated the performance of Bitnet.cpp for end-to-end edge inference for ternary LLM. Compared to state-of-the-art methods, Bitnet.cpp significantly improves ternary LLM edge inference performance across different CPU architectures and model sizes under the sub-2-bits-per-weight condition. For quality evaluation, compared to Float16, TL1\_0 and TL2\_0 exhibit negligible loss, whereas I2\_S, TL1\_1, and TL2\_1 achieve lossless in BitNet b1.58.


\subsection{Speed Evaluation}
\subsubsection{Devices}
We conducted a performance evaluation of Bitnet.cpp on two devices: the Apple M2 Ultra and the Intel i7-13700H. These devices represent the ARM and x86 architectures, respectively, covering most edge devices and ensuring broad applicability and reliable performance results for Bitnet.cpp.

\subsubsection{Baselines}
We conducted experiments from two perspectives: lossless inference and fast inference. For the lossless inference aspect, we chose llama.cpp Float16 as the baseline and compared it with I2\_S from Bitnet.cpp. This comparison evaluates the lossless inference performance of Bitnet.cpp, demonstrating its improvements in both accuracy and speed. For the fast inference aspect, we conducted experiments based on the two features of TL2\_0: element-wise and LUT-based. llama.cpp includes two element-wise MAD-based solutions, TQ1\_0 and TQ2\_0. To neutralize the effect of bpw, TQ1\_0, which has a bpw nearly identical to TL2\_0, was selected for comparison. This comparison aims to evaluate the performance differences between MAD-based and LUT-based solutions. For T-MAC, a bit-wise LUT-based solution, the 2-bit kernel was selected for comparison with TL2\_0 to assess performance differences between element-wise and bit-wise methods.

\subsubsection{End-to-end Inference Speed}
We evaluated the token generation speed of Bitnet.cpp and observed a significant speed advantage across different CPU architectures and model sizes compared to baselines. As illustrated in Figure~\ref{fig:whole-compa}, I2\_S achieves up to a 6.25x speedup compared to Float16, demonstrating that Bitnet.cpp provides a comprehensive advantage in both accuracy and speed. Furthermore, TL2\_0 outperforms T-MAC by up to 2.32x on the Intel i7-13700H and by up to 1.19x on the Apple M2 Ultra, indicating a notable improvement in LUT-based mpGEMM performance. Moreover, TL2\_0 surpasses TQ1\_0, with up to 1.33x speedup on the Intel i7-13700H and 1.65x on the Apple M2 Ultra, further improving performance in element-wise mpGEMM with bpw below 2. As detailed in Table~\ref{tab:detailed_com}, TL2\_0 reaches 7.45 tokens/s on the Apple M2 Ultra and 1.69 tokens/s on the Intel i7-13700H, outperforming previous ternary kernels in 100B ternary LLM inference on edge devices. These findings highlight the significant inference benefits of Bitnet.cpp.

\subsection{Quality Evaluation}
We used the bitnet\_b1\_58-large\footnote{\url{https://huggingface.co/1bitLLM/bitnet_b1_58-large}} model and the perplexity\footnote{\url{https://github.com/ggerganov/llama.cpp/tree/master/examples/perplexity}} tool from llama.cpp for quality evaluation. For baselines, Float16 and Q4\_0 from llama.cpp were selected for comparison with Bitnet.cpp. For tasks, we used WikiText2\cite{wikitext2} to measure perplexity (the lower, the better), HellaSwag\cite{hellaswag} and WinoGrande\cite{winograde} to measure accuracy (the higher, the better). As shown in Table~\ref{tab:accuracy}, both TL1\_0 and TL2\_0 achieve nearly identical perplexity compared to Float16 on WikiText2 and maintain accuracy comparable to Float16 on WinoGrande and HellaSwag. I2\_S, TL1\_1, and TL2\_1 exhibit lossless performance relative to Float16 across all tasks. These results indicate that the loss introduced by Bitnet.cpp is negligible.

\begin{table}[htbp]
    \centering
    \footnotesize
    \captionsetup{skip=8pt}
    \renewcommand{\arraystretch}{1.4}
    \begin{tabular}{|p{1cm}|p{1.3cm}|p{1.3cm}|p{1.3cm}|}
        \hline
        \multirow{2}{*}{Method} 
        & \textbf{WikiText2} & \textbf{Winograd} & \textbf{HellaSwag} \\
        & Perplexity\(\downarrow\) & Accuracy\(\uparrow\)  & Accuracy\(\uparrow\) \\
        \hline
        Float16& 11.29 & 55.32 & 43.0 \\
        \hline
        Q4\_0  & 11.57 & 55.09 & 42.25 \\
        \hline
        TL1\_0 & 11.30 & 55.32 & 43.0 \\
        \hline
        TL2\_0 & 11.30 & 55.32 & 43.0 \\
        \hline
        TL1\_1 & 11.29 & 55.32 & 43.0 \\
        \hline
        TL2\_1 & 11.29 & 55.32 & 43.0 \\
        \hline
        I2\_S & 11.29 & 55.32 & 43.0 \\
        \hline
    \end{tabular}
    \caption{\label{tab:accuracy}End-to-end inference quality.}
\end{table}

\section{Related Work}
\textbf{LUT-based mpGEMM}
Previous research has explored the application of LUT-based mpGEMM in deep learning. \cite{deepgemm} employs LUT-based mpGEMM to accelerate computations in convolutional neural networks, while \cite{MADDNESS, LUT-NN} utilize this approach to process vector-quantized activations. For LLM inference, \cite{lut-gemm, gpu-lut1} apply LUT-based GEMM on GPUs. However, in practice, these methods are often slower than MAD-based approaches, such as \cite{cutlass, bitblas}, due to the inefficiency of rapid table access on GPU.

\noindent \textbf{LLM Inference}
FlashAttention \citep{flashatten, flashatten2} introduces an innovative approach to GPU attention kernel design. VLLM \citep{vLLM} and TensorRT-LLM \citep{trt-LLM} have optimized end-to-end inference performance using systematic techniques. Powerinfer \cite{powerinfer, powerinfer2} proposes novel strategies that intelligently balance workloads across heterogeneous devices, improving overall inference efficiency.

\noindent \textbf{LLM Quantization}
Post-training quantization (PTQ) refers to converting a full-precision LLM to a low-precision without retraining, with related works including \cite{smoothquant, AWQ, quip, gptq, spqr, llmint8, omniquant}. However, PTQ inevitably results in quantization loss. In contrast, Quantization-Aware Training (QAT) effectively avoids this issue. QAT involves retraining a pretrained model to obtain a quantized model, thus mitigating quantization loss. Relevant works include \cite{liu2023llmqatdatafreequantizationaware, eff-qat, bitdistill}. BitNet B1.58 adopts QAT, creating conditions for lossless inference in the system.

\section{Conclusion}
In this paper, by optimizing mpGEMM, we address the inefficiencies caused by the conflicts of non-integer bpw in ternary LLMs with memory access alignment rules, and enable lossless inference for BitNet b1.58. Our key idea is to utilize a finer-grained element-wise scheme instead of bit-wise, and consistent with BitNet b1.58 training schemes. Based on our key ideas, we develop Bitnet.cpp, featuring TL, the first element-wise LUT-based mpGEMM kernel for ternary LLMs, and I2\_S, the first lossless MAD-based kernel for BitNet b1.58. The practical outcomes of our research are noteworthy. We have demonstrated that Bitnet.cpp achieves up to 6.25x speedup compared to baselines and provided lossless inference for BitNet b1.58. To enhance the generality of our research, we extended the TL to ELUT for low-bit LLMs, highlighting its efficiency and potential. This paper presents extensive work on optimizing edge inference for ternary LLMs from both algorithmic and engineering perspectives. It offers the research community new insights into handling ternary and non-integer bpw weights, shows the practical advantages of ternary LLMs and presents the industry with innovative solutions for deploying fast, lossless LLMs on edge devices.

\newpage

\section*{Limitations} 
Bitnet.cpp has the following limitations:
\begin{itemize}[left=0pt]
\item
Bitnet.cpp currently only provides a practical solution for ternary LLM inference on edge devices. In the future, we plan to extend the Bitnet.cpp to offer efficient inference solutions for ternary LLMs across multiple devices.
\item 
Bitnet.cpp is specifically designed for ternary LLMs, with a relatively narrow range of applicable model architectures. In response to this, we have expanded the element-wise LUT-based (ELUT) method to cover low-bit ranges in the appendix. However, it still lacks support from actual LLMs other than ternary ones.
\item
Bitnet.cpp does not discuss in detail the acceleration specifics of LLMs during the prefilling stage, as there has been a shift in the resource bottleneck from being memory-bound during the decoding stage to computation-bound during the prefilling stage. Therefore, the original optimization methods are no longer applicable, and we will continue to explore optimization methods for the prefilling stage.
\end{itemize}


\newpage

\appendix
In the appendix, we extend the concept of element-wise LUT-based solutions beyond ternary LLMs, analyzing its capabilities and potential from a more general perspective.

\section{\label{sec:insight}Insight}
In this section, we will analyze the computational complexity and memory access complexity of the element-wise LUT-based (ELUT) mpGEMM algorithm. Based on this analysis, we will compare our results with those of MAD-based solutions and bit-wise LUT-based solutions, drawing the conclusion that the ELUT algorithm exhibits comprehensive advantages in both computation and memory access compared to previous algorithms.

\subsection{Complexity}

In general, mpGEMM requires two steps to complete: the preprocessing stage and the accumulation stage. As shown in Algorithm \ref{alg:GEN-MAD}, for the MAD-based solution, the preprocessing stage involves quantizing the floating-point activations to integers, with a computational complexity of $O(NK)$ and a memory access complexity of $O(NK)$. In the accumulation stage, the MAD-based solution performs element-wise multiplication and accumulation for the K corresponding elements across M rows and N columns, resulting in a computational complexity of $O(MNK)$ and a memory access complexity of $O(MNK)$.

As shown in Algorithm \ref{alg:GEN-LUT}, for ELUT, the preprocessing stage involves first performing quantization to quantize the floating-point activations into $NK/g$ groups, and then enumerating the $C^g$ possible values within each group to construct the Lookup Table. The computational complexity of this process is $O(NKC^g/g)$, and the memory access complexity is also $O(NKC^g/g)$. In the accumulation stage, ELUT performs lookup and accumulation operations group by group. The computational complexity of this process is $O(MNK/g)$, while the memory access complexity is $O(MNKC^g/g)$ because the entire Lookup Table must be loaded for each group.

\begin{algorithm}
\SetAlgoLined
\DontPrintSemicolon
\caption{MAD-based mpGEMM}
\label{alg:GEN-MAD}
\KwIn{Activation \( A \) of shape \( N, K \)}
\KwIn{Weights \( W \) of shape \( M, K \), \( W \in \mathbb{Z}, \ |W| = C \)}
\KwOut{Result matrix \( R \) of shape \( M, N \)}
\tcc{C-complexity $\rightarrow$ Computational Complexity}
\tcc{M-complexity $\rightarrow$ Memory Access Complexity}
\tcc{Phase1 : Preprocessing}
\tcc{C-complexity : $O(NK)$ / M-complexity : $O(NK)$}
\( A_q = \text{Quantization}(A) \) \\
\tcc{Phase2 : Accumulation}
\tcc{C-complexity : $O(MNK)$ / M-complexity : $O(MNK)$}
\For{$ m,n \gets 1 \text{ to } M,N $}
    {
        \( R[n, m] = \sum_{k=1}^{K} (A_q[n, k] * W[m, k]) \)
    }
\tcc{Overall C-complexity : $O(MNK)$} 
\tcc{Overall M-complexity : $O(MNK)$}
\end{algorithm}

\begin{algorithm}
\SetAlgoLined
\DontPrintSemicolon
\caption{ELUT mpGEMM}
\label{alg:GEN-LUT}
\KwIn{Activation \( A \) of shape \( N, K \)}
\KwIn{Weights \( W \) of shape \( M, K \), \( W \in \mathbb{Z}, \ |W| = C \)}
\KwIn{Group size \( g \)}
\KwOut{Result matrix \( R \) of shape \( M, N \)}
\tcc{C-complexity $\rightarrow$ Computational Complexity} 
\tcc{M-complexity $\rightarrow$ Memory Access Complexity} 
\tcc{Phase1 : Preprocessing}
\tcc{C-complexity : $O(NKC^g/g)$ / M-complexity : $O(NKC^g/g)$}
\( A_q = \text{Tbl-quantization}(A) \) \\
\( LUT_A = \text{Table-setup}(A_q) \) \\
\tcc{Phase2 : Accumulation}
\tcc{C-complexity : $O(MNK/g)$ / M-complexity : $O(MNKC^g/g)$}
\For{$ m,n \gets 1 \text{ to } M,N $}
    {
        \( R[n, m] = \sum_{k=1}^{K/g} \text{Lookup}(LUT_A[n, k], W[m, k]) \)
    }
\tcc{Overall C-complexity : $max(O(NKC^g/g),O(MNK/g))$} 
\tcc{Overall M-complexity : $O(MNKC^g/g)$}
\end{algorithm}

Through theoretical analysis, we can identify several interesting insights. First, ELUT has an advantage over the MAD-based solution in terms of computational complexity for LLM inference. The overall computational complexity of the MAD-based solution is $O(MNK)$, while ELUT is $max(O(NKC^g/g),O(MNK/g))$. This implies that as long as $C^g < M$ and $g > 1$, ELUT requires fewer computations for mpGEMM. In LLMs, the value of $M$, i.e., the hidden size, is generally large. In contrast, the $C$ value for ternary LLMs is only 3 and $g$ is only 2 or 3. Therefore, ELUT is computationally more efficient than the MAD-based solution.

However, ELUT has a disadvantage in terms of memory access complexity compared to the MAD-based solution. The memory access complexity of the MAD-based solution is $O(MNK)$, while the LUT-based solution has a memory access complexity of $O(MNKC^g/g)$. In practical implementations, we employ optimization techniques such as element-wise mirror consolidation and LUT-centric data layout to reduce memory access complexity, thereby significantly mitigating the overhead caused by memory access.

\begin{figure*}[ht]
    \centering
    \includegraphics[width=\linewidth]{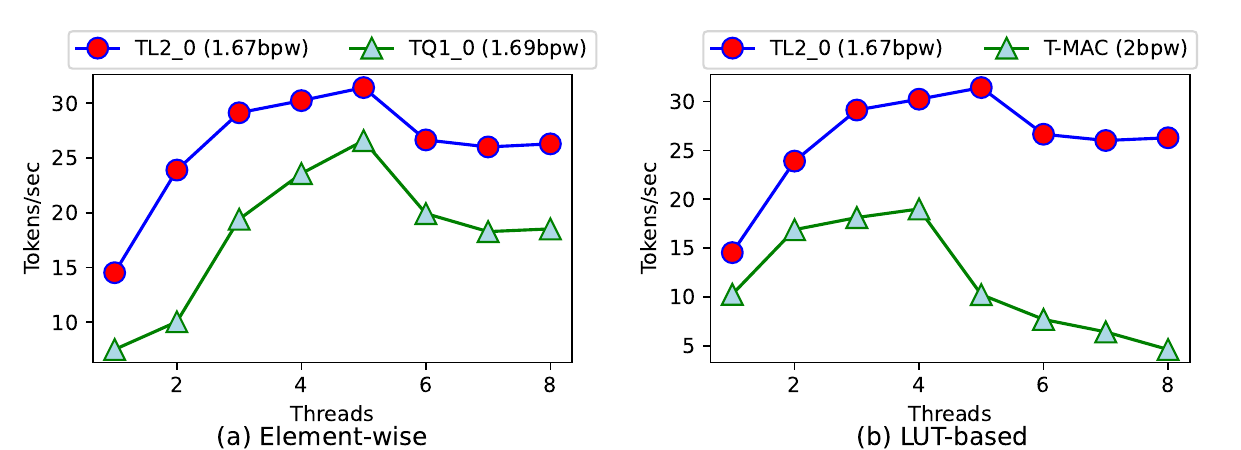}
    \caption{Multi-threaded end-to-end inference performance of the 3.8B model on Intel i7 13700H.}
    \label{fig:ana-plot}
\end{figure*}

\subsection{Compared to MAD-based: More Practical}
In fact, when deploying LLMs on current edge devices, we often face the limitation of using only a very small number of threads. Under such circumstances, the constraints on computational resources are maximized, making computational complexity a critical factor. In contrast, due to the limited number of threads, memory access is unlikely to reach bandwidth limits. In this context, ELUT, with its computational complexity being only $\frac{1}{g}$ of that of the MAD-based solution in most cases, is expected to outperform the MAD-based solution in real-world inference scenarios for LLMs. Therefore, ELUT is more suitable for deployment in practical scenarios than the MAD-based solution.

\subsection{Compared to Bit-Wise : More Fine-grained}

\begin{table}[ht]
\centering
\begin{tabular}{|c|c|c|c|}
\hline
$C$ & $g$ & $bpw_b$ & $bpw_e$\\
\hline
3 & 3 & 2 & \textbf{1.67} \\
\hline
4 & 2 & 2 & 2\\
\hline
5 & 2 & 3 & \textbf{2.5}\\
\hline
...& ... & ... & ...\\

\hline
\end{tabular}
\caption{A comparison table of bpw from bit-wise and element-wise for different weight cardinality. $C$ represents the weight cardinality, $g$ indicates to group size, $bpw_b$ denotes bit-wise bpw, $bpw_e$ refers to element-wise bpw.}
\label{tab:valuewise}
\end{table}

Although we have demonstrated that ELUT outperforms MAD-based solutions in terms of performance with low thread counts, the bit-wise LUT-based solution also exhibits this advantage. The advantage of the ELUT method over the bit-wise method lies in its finer granularity of LUTs, shifting from bit-based to element-based, ensuring a more information-preserving compression of weights.

Returning to the computational complexity, in most cases, the computational complexity of the LUT method is \(O(MNK/g)\). For ternary LLMs, when \(g = 3\), the complexity is reduced by a factor of $\frac{1}{6}$ compared to \(g = 2\). In terms of memory access complexity, since mirror consolidation is used when \(g = 3\), we can compute the memory access complexity for \(g = 2\) and \(g = 3\) as follows.

\[
 O(\frac{MNK3^2}{2}) = O(\frac{MNK3^3/2}{3})
\]

Based on this, since the \(bpw\) when \(g = 3\) is approximately 1/6 lower than when \(g = 2\) and memory access complexity is similar, we observe that when using the ELUT method on ternary LLMs inference, both computation and memory access are reduced compared to the bit-wise method. Similarly, as Table \ref{tab:valuewise} shown, the same conclusion can be extended to the case where \(C \neq 2^n\). This provides theoretical guidance for TL implementation.

\section{\label{sec:analy}Analysis}
\subsection{Memory-Computation Trade-off Decoding}
During the execution of a kernel, the execution speed is determined by both instruction computation speed and data access speed. The instruction computation speed is related to the computational complexity, instruction types, and the depth of the pipeline, while the data access speed depends on the memory access complexity, locality, and the type of memory being accessed. The kernel execution speed is ultimately determined by the smaller of these two values. Naturally, we refer to computation-related consumptions as \textbf{computation} consumptions and data-access-related consumptions as \textbf{memory} consumptions. Thus, optimizing kernel performance is essentially a process of exploring the compute-memory trade-off. In fact, ELUT outperforms previous approaches in achieving a better trade-off, resulting in performance improvements. This can be clearly observed from both the compute and memory perspectives by analyzing performance gap for TQ1\_0 and T-MAC with TL2\_0.

\subsection{Towards Memory: Compared to T-MAC}
In the previous section, we provided a detailed theoretical analysis of the LUT-based solution, showing that the memory access complexity of ELUT and T-MAC is equivalent, but with a lower bpw, resulting in reduced memory access requirements. In the following, we validate this conclusion with practical examples.

In fact, TL2\_0 has an advantage over T-MAC in terms of bpw, which enhances the performance ceiling of memory-intensive LUT-based solutions to some extent. As a result, significant performance improvements are observed, particularly in low bandwidth environments. As shown in Figure \ref{fig:ana-plot} (b), TL2\_0 achieves a performance improvement over T-MAC in a multi-threaded environment. Notably, the performance of TL2\_0 continues to improve as the number of threads reaches 5, while the speed of T-MAC begins to decline. This indicates that TL2\_0 reaches the memory-bound state later than T-MAC, thereby raising the performance ceiling.

\subsection{Towards Compute: Compared to TQ1\_0}
In the previous section, we theoretically verified that ELUT exhibits lower computational complexity compared to the MAD-based solution. To ensure a fair comparison, we selected TQ1\_0, which has a bpw almost identical to that of TL2\_0, for the comparative experiment. The results show that LUT-based solutions offer an advantage over MAD-based solutions in terms of computation-related consumption, leading to a significant performance improvement. As shown in Figure \ref{fig:ana-plot} (a), the shape of performance curves of TL2\_0 and TQ1\_0 in a multi-threaded environment are nearly identical, with TL2\_0 consistently outperforming TQ1\_0 across all threads. This further supports our conclusion that LUT-based solutions have an advantage over MAD-based solutions in computation-related consumption, resulting in a significant performance increase.

\begin{figure*}[htbp]
    \centering
    \begin{minipage}{0.45\textwidth}
        \centering
        \includegraphics[width=\columnwidth]{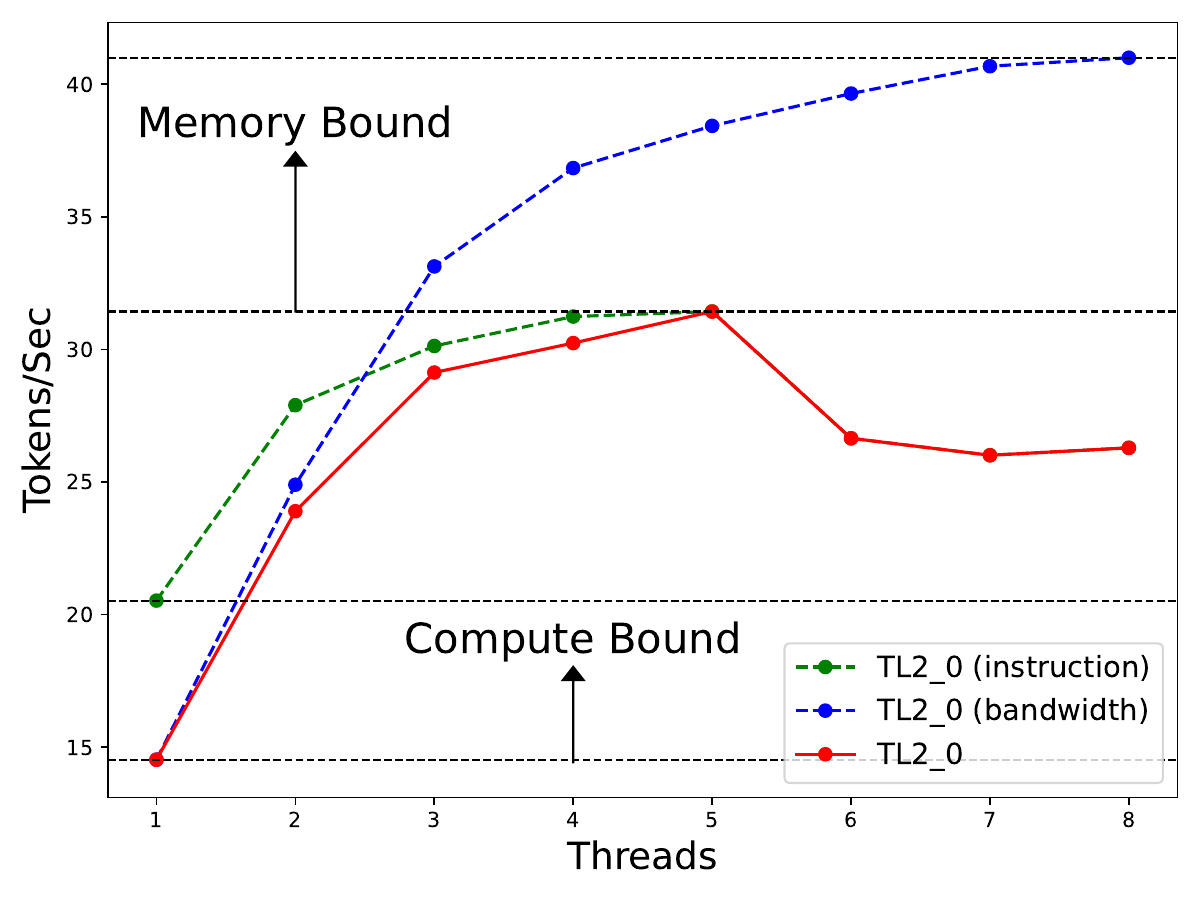}
        \caption{\label{fig:potential}ELUT performance potential curve.}
    \end{minipage}\hfill
    \begin{minipage}{0.45\textwidth}
        \centering
        \includegraphics[width=\columnwidth]{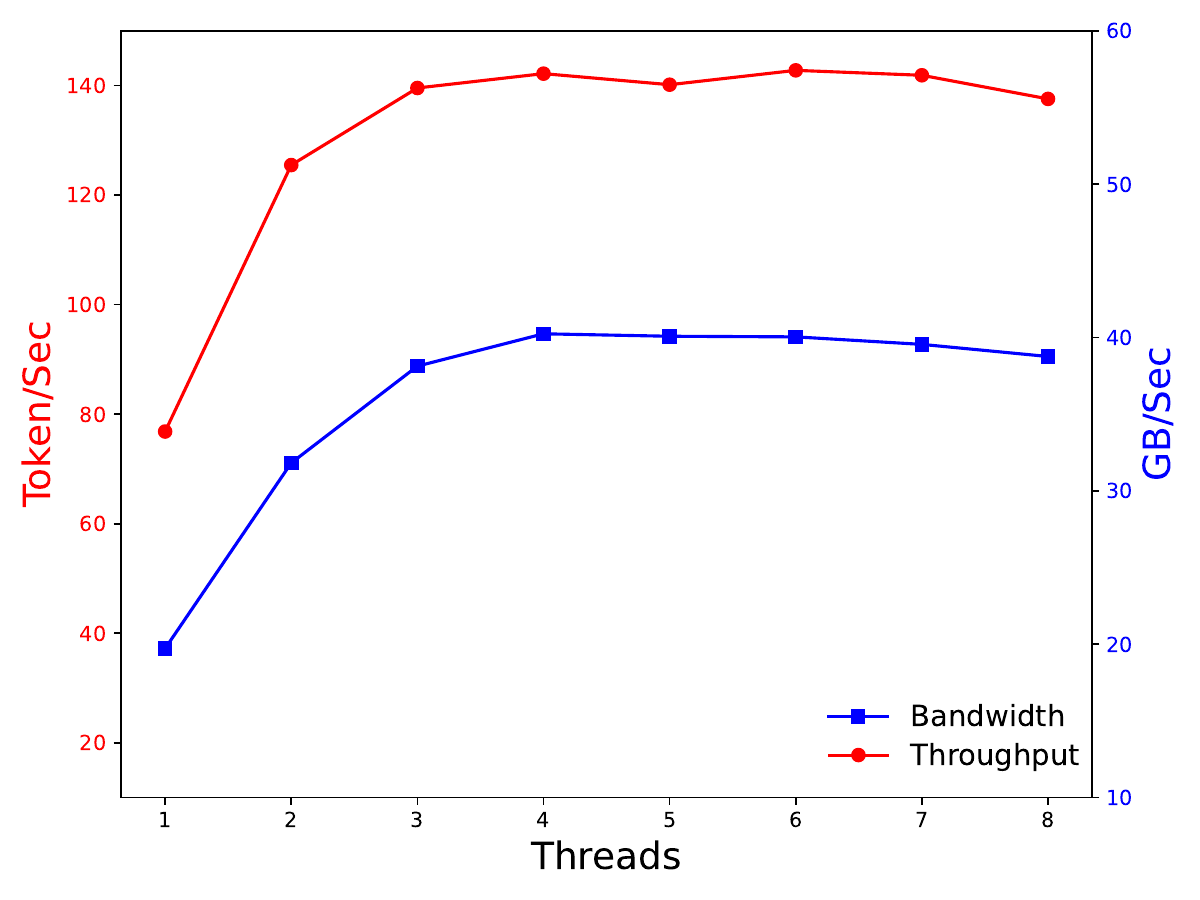}
        \caption{\label{fig:twin}Throughput and Bandwidth curve, tested with bitnet-b1.58-large on intel core i5-13400F.}
    \end{minipage}
\end{figure*}

\section{\label{sec:poten}Potential}
After evaluating the performance of ELUT, we have observed that it has a comprehensive advantage over other methods. However, we believe that ELUT has not yet reached its theoretical performance limit. In the following, we will analyze the hardware limitations affecting ELUT and estimate its theoretical performance in the absence of such constraints. This analysis aims to explore the potential of ELUT and provide insights for future hardware designs targeting low-bit LLMs inference.

\subsection{Bandwidth}

Bandwidth is the data transfer rate between memory and the processor, and it also determines the execution rate of kernels. Considering that ELUT has a higher memory access complexity than the MAD-based solution, bandwidth has a significant influence on overall end-to-end inference speed. As shown in Figure \ref{fig:whole-compa}, it is evident that TL2\_0 demonstrates a more pronounced acceleration effect on T-MAC for Intel i7-13700H compared to Apple M2 Ultra. The main reason for this phenomenon lies in the significant difference in maximum bandwidth between the two edge devices. In fact, the Apple M2 Ultra has a maximum bandwidth exceeding 800 GB/s, while the maximum bandwidth of the Intel i7-13700H is less than 100 GB/s. As shown in Figure \ref{fig:twin}, we used PCM \cite{PCM} tool to measure the token throughput and bandwidth at different thread counts and compared them side by side. It is clear that the shape of token throughput and bandwidth curves are nearly identical. When the thread count reaches 4, the token throughput also reaches its maximum value due to the saturation of the bandwidth, causing the end-to-end inference speed to reach its peak. Therefore, we can conclude that the maximum bandwidth limits the potential of ELUT. Building on this, as shown in Figure \ref{fig:potential}, we estimated the end-to-end inference speed when the bandwidth is increased. We anticipate that, with the increase in maximum bandwidth, ELUT will reach the memory-bound state later, resulting in a higher end-to-end inference speed, with the upper bound still determined by the theoretical maximum bandwidth. This estimation validates our theoretical analysis of ELUT. Moreover, we are pleased to note that there is currently a trend towards increasing the bandwidth of edge devices, which will further unlock the potential of ELUT.

\begin{table*}[ht]
\centering
\begin{tabular}{cccc}
\hline
Instruction Set & LUT-based & MAD-based & \\
\hline
AVX2& \_mm256\_shuffle\_epi8 & \_mm256\_maddubs\_epi16&\\
NEON&  vqtbl1q\_u8 & vmlal\_s8 / vmull\_s16 + vaddq\_s32&\\
\hline
\end{tabular}
\caption{\label{tab:wise-instruct}Core instructions in AVX2 and Neon for LUT-based and MAD-based mpGEMM.}
\end{table*}

\subsection{Instructions Throughput}


SIMD instructions are commonly used to implement kernels on CPUs, as SIMD allows a single instruction to process multiple data elements simultaneously, achieving computation parallelism and acceleration. 
For SIMD instructions, two metrics determine the performance of the instruction: instruction throughput, which determines the number of instructions that can be completed in a single clock, and instruction latency, which determines the number of clocks required to complete a single instruction. On modern CPUs, since MAD operations are widely used, common architectures such as x86 and ARM have made specific optimizations to ensure high instruction throughput for these operations (as shown in Table \ref{tab:wise-instruct}). For example, in the x86 architecture with AVX2 instructions, a single MAD instruction can complete an int8 multiply-accumulate operation and convert the result to int16. However, for ELUT, we need to use three types of instructions—TBL (table lookup), ADD (accumulation), and CVT (type conversion)—to accomplish the same task. Although the AVX documentation \footnote{\url{https://www.intel.com/content/www/us/en/docs/intrinsics-guide/index.html}} states that the latency of the MAD instruction is 5 cycles, which is greater than the latency of the TBL instruction, both instructions have the same throughput. This implies that, under reasonable pipeline scheduling, the theoretical completion time for MAD and TBL instructions is the same. We validated this on an Intel i5-13400F, where the completion time for a single MAD instruction was \textbf{3.77 ns}, and for a single TBL instruction, it was \textbf{3.70 ns}, which is nearly identical. However, since the table lookup must be followed by addition and conversion (TBL+ADD+CVT), this sequence inevitably leads to a reduction in throughput. We observed that completing the same task with TBL+ADD+CVT took \textbf{6.20 ns}, approximately 68\% longer than the raw latency of a single MAD instruction. This highlights that, in terms of throughput, the table lookup followed by the accumulation method suffers significant performance loss due to insufficient hardware support.

In previous work, \cite{mo2024luttensorcorelookup, LUTMUL} was implemented in hardware on GPUs and FPGAs, respectively, as solutions similar to ELUT, and they achieved performance improvements over MAD-based solutions. This suggests that providing better hardware support for ELUT on edge devices is highly promising. As shown in Figure \ref{fig:potential}, we estimated the performance of ELUT with hardware support, and the results indicate a significant performance boost when bandwidth is not a bottleneck. We sincerely hope that the exploration of ELUT's potential can inspire future hardware designs to fully unlock ELUT's capabilities.

\subsection{Register Length}

\begin{figure}[ht]
    \centering
    \includegraphics[width=\columnwidth]{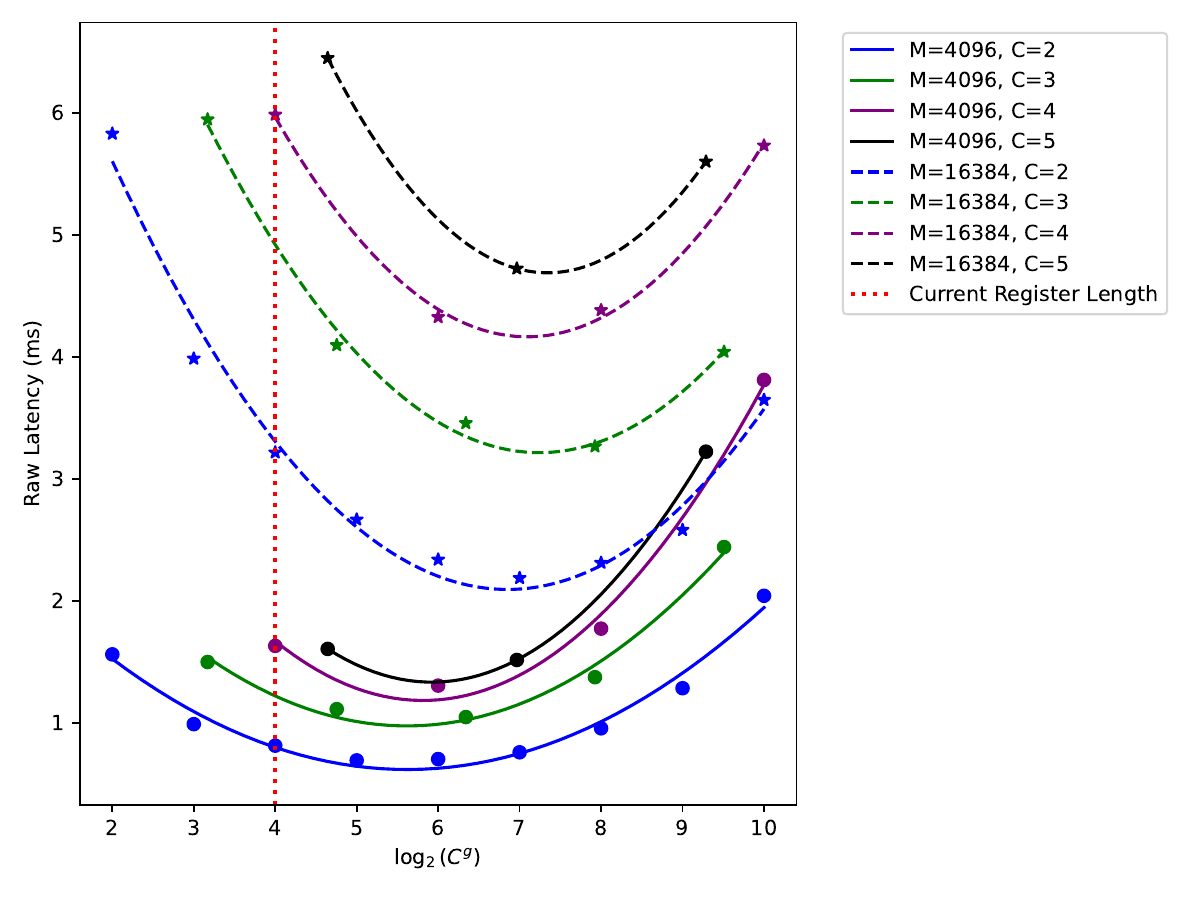}
    \caption{\label{fig:reg} Register length and raw latency relationship graph.}
\end{figure}

The length of registers also imposes a limitation on the performance of ELUT. Taking AVX2 as an example, the lookup width of the TBL SIMD instruction is 128 bits, which means that it can look up 16 int8 values in one operation. Clearly, from an element-wise perspective, all the possible values of \( C^g \) that we enumerate need to be covered in a single lookup. Otherwise, we would need to use a bit-wise approach, performing bit-by-bit lookups, which sacrifices the memory access benefits obtained from the element-wise method. For example, in the case of ternary LLMs, with the limitation of 128-bit register length, we can enumerate at most \( \frac{3^3}{2} \) possible values in the lookup table, which restricts \( g \leq 3 \). Assuming we disregard the limitation of instruction length, we simulate a longer instruction length using the original instructions without considering precision. As shown in Figure \ref{fig:reg}, as the length of SIMD registers increases, the number of enumerable \( g \) values grows, thereby significantly reducing computational complexity. Theoretically, when \( C^g = M \), the computational complexity introduced by enumerating LUTs surpasses that of table lookup and accumulation, and further increasing the length of SIMD registers no longer yields additional benefits. It is significant that the \( g \) values we can currently enumerate are still far from the intersection point. Therefore, increasing the register length provides a definite benefit in terms of computational complexity. This also indicates that the potential of ELUT has not yet reached its theoretical limit.


\newpage
\onecolumn

\section{TL Algorithm}

\begin{table}[H]
\centering
\begin{tabular}{|>{\centering\arraybackslash}p{1cm}|>{\centering\arraybackslash}p{1cm}|>{\centering\arraybackslash}p{1.5cm}|}
\hline
\multicolumn{2}{|c|}{\textbf{Unpack}} & \textbf{Pack} \\ \hline
-1 & -1 & 0000 \\ \hline
-1 & 0 & 0001 \\ \hline
-1 & 1 & 0010 \\ \hline
0 & -1 & 0011 \\ \hline
0 & 0 & 0100 \\ \hline
0 & 1 & 0101 \\ \hline
1 & -1 & 0110 \\ \hline
1 & 0 & 0111 \\ \hline
1 & 1 & 1000 \\ \hline
\end{tabular}
\vspace{0.3cm}
    \caption{TL1 Kernel transforms every two full-precision weights into 4-bit index and performs LUT computation.}
    \label{tab:tl1}
\end{table}

\begin{algorithm}
\SetAlgoLined
\DontPrintSemicolon
\caption{TL1 mpGEMM}
\label{alg:TL1}
\KwIn{Activation \( A \) of shape \( N, K \)}
\KwIn{Weights \( W \) of shape \( M, K \)}
\KwOut{Result matrix \( R \) of shape \( M, N \)}

\SetKwFunction{FPW}{PreprocessWeights}
\SetKwFunction{FPC}{PreCompute}
\SetKwProg{Fn}{Function}{:}{}

IndexWeight = \FPW{W, M, K}

LUT = \FPC{A, N, K}

\For{$ n, m \gets 1 \text{ to } N, M $}
    { 
        \( R[n, m] = \sum_{k = 1}^{K/2} \text{Lookup}(LUT, IndexWeight, n, m, k) \)
    }

\Fn{\FPC{$A, N, K$}}{
    \For{$ n, k \gets 1 \text{ to } N, K/2 $}
        {
            \For{$ i \gets 1 \text{ to } 3^2$}
            {
            \tcc{\( Unpack\) shows in Table \ref{tab:tl1}}
                \( LUT[n, k, i] = Unpack_i(A[n, 2k], A[n, 2k+1]) \)
            }
        }
    \textbf{return} \( R \)
}

\Fn{\FPW{$W, M, K$}}{
    \For{$m, k \gets 1 \text{ to } M, K / 2$}
        {
            \tcc{\( Pack\) shows in Table \ref{tab:tl1}}
            \( IndexWeight[m, k] = Pack(W[m, 2k], W[m, 2k+1]) \)
        }
    \KwRet{$IndexWeight$}
}
\end{algorithm}

\newpage

\begin{table}[H]
\centering
\begin{tabular}{|>{\centering\arraybackslash}p{1cm}|>{\centering\arraybackslash}p{1cm}|>{\centering\arraybackslash}p{1cm}|>{\centering\arraybackslash}p{1.5cm}|}
\hline
\multicolumn{3}{|c|}{\textbf{Unpack}} & \textbf{Pack} \\ \hline
-1 & -1 & -1 & \textcolor{red}{1}\space1101 \\ \hline
-1 & -1 & 0 & \textcolor{red}{1}\space1100 \\ \hline
-1 & -1 & 1 & \textcolor{red}{1}\space1011 \\ \hline
-1 & 0 & -1 & \textcolor{red}{1}\space1010 \\ \hline
\multicolumn{4}{|c|}{...}\\
\hline
0 & 0 & 0 & \textcolor{red}{0}\space0000
 \\ \hline
\multicolumn{4}{|c|}{...}
\\ \hline
1 & 0 & 1 &  \textcolor{red}{0}\space1010\\ \hline
1 & 1 & -1 & \textcolor{red}{0}\space1011 \\ \hline
1 & 1 & 0 & \textcolor{red}{0}\space1100 \\ \hline
1 & 1 & 1 & \textcolor{red}{0}\space1101 \\ \hline
\end{tabular}
\vspace{0.3cm}
    \caption{TL2 Kernel compresses every three full-precision weights into a 1-bit sign (\textcolor{red}{0} or  \textcolor{red}{1}) and a 4-bit index. }
    \label{tab:tl2}
\end{table}

\begin{algorithm}
\SetAlgoLined
\DontPrintSemicolon
\caption{TL2 mpGEMM}
\label{alg:TL2}
\KwIn{Activation \( A \) of shape \( N, K \)}
\KwIn{Weights \( W \) of shape \( M, K \)}
\KwOut{Result matrix \( R \) of shape \( M, N \)}

\SetKwFunction{FPW}{PreprocessWeights}
\SetKwFunction{FPC}{PreCompute}
\SetKwProg{Fn}{Function}{:}{}

IndexWeight, Signweight = \FPW{W, M, K}

LUT = \FPC{A, N, K}

\For{$ n, m \gets 1 \text{ to } N, M $}
    {
        \( R[n, m] = \sum_{k \gets 1}^{K/3} \text{Lookup}(LUT, IndexWeight, n, m, k) \) \\
        \( R[n, m] = Signweight \times R[n, m] \)
    }

\Fn{\FPC{$A, N, K$}}{
    \For{$ n, k \gets 1 \text{ to } N, K/3 $}
        {
            \For{$ i \gets 1 \text{ to } 3^3 / 2$}
            {
            \tcc{\( Unpack\) shows in Table \ref{tab:tl2}}
                \( LUT[n, k, i] = Unpack_i(A[n, 3k], A[n, 3k+1], A[n, 3k+2]) \)
            }
        }
    \textbf{return} \( R \)
}

\Fn{\FPW{$W, M, K$}}{
    \( SignWeight = Sign(W) \)

    \( W = \left| W \right| \)

    \For{$ m, k \gets 1 \text{ to } M, K/3 $}
        {
            \tcc{\( Pack\) shows in Table \ref{tab:tl2}}
            \( IndexWeight[m, k] = Pack(W[m, 3k], W[m, 3k+1], W[m, 3k+2]) \)
        }
    \KwRet{$IndexWeight, SignWeight$}
}
\end{algorithm}

\newpage
\section{Performance}

\begin{table*}[htbp]
    \centering
    \renewcommand{\arraystretch}{1.4}
    \resizebox{\textwidth}{!}{
    \begin{tabular}{|l|l|c|c|c|c|c|c|c|c|c|c}
        \hline
        \multirow{4}{*}{\makecell[l]{CPU}} 
        & \multirow{4}{*}{\makecell[l]{Model \\ Size}} & \multicolumn{8}{c|}{Kernels} \\
        \cline{3-10}
        &                    & \multicolumn{3}{c|}{general kernels} & \multicolumn{5}{c|}{ternary kernels} \\
        \cline{3-10}
        &                    & Float16 & Q4\_0 & T-MAC & TQ1\_0 & TQ2\_0 & TL1\_0 & TL2\_0 & I2\_S\\
        &                    & b(16) & b(4.5) & b(2) & b(1.69) & b(2.06) & b(2) & \textbf{b(1.67)} & b(2)\\
        \hline
        \multirow{9}{*}{\makecell[l]{Intel i7-13700H \\ 20C 64G}}
        & 700M & 30.73 & 67.57 & 76.29 & 114.20& 123.94& 75.62 & \textbf{126.99} & 125.37\\
        & 1.5B & 15.02 & 35.46 & 42.38 & 64.86 & 71.92 & 43.44 & 74.16 & \textbf{77.75}\\
        & 3.8B & 5.85  & 16.33 & 18.12 & 26.59 & 33.19 & 17.91 & \textbf{35.43} & 35.04\\
        & 7B   & 3.30  & 9.09  & 12.29 & 17.96 & 19.92 & 11.89 & \textbf{20.72} & 20.62\\
        & 13B  & 1.78  & 5.04  & 6.44  & 10.55 & 11.21 & 6.32  & \textbf{11.41} & 10.62\\
        & 30B  & N/A   & 2.13  & 2.54  & 4.62  & 5.25  & 2.65  & 4.99 & \textbf{5.70}\\
        & 70B  & N/A   & 0.94  & 1.32  & 2.09  & 2.32  & 1.49  & \textbf{2.42} & 2.30\\
        & 100B & N/A   & 0.67  & 0.73  & 1.48  & 1.61  & 0.75  & \textbf{1.69} & 1.65\\
        \hline
        \multirow{9}{*}{APPLE M2} 
        & 700M & 110.65& 197.38& 220.22& 217.64& 237.61 & 214.53& 229.21 & \textbf{238.16}\\
        & 1.5B & 59.49 & 117.77& 135.27& 130.10& \textbf{145.68} & 132.68& 138.28 & 143.43\\
        & 3.8B & 28.31 & 71.89 & 91.84 & 73.14 & 88.66  & 90.73 & \textbf{92.12} & 91.65\\
        & 7B   & 14.87 & 39.47 & 53.37 & 45.55 & 54.90  & 52.77 & \textbf{55.42} & 54.74\\
        & 13B  & 8.42  & 23.28 & 31.72 & 25.83 & \textbf{34.63}  & 32.12 & 33.22 & 32 .88\\
        & 30B  & 3.78  & 10.98 & 16.40 & 12.85 & 15.46  & 15.02 & \textbf{19.59} & 16.41\\
        & 70B  & 1.71  & 5.16  & 9.48  & 6.30  & 8.16   & 9.23  & \textbf{10.37} & 8.39\\
        & 100B & N/A   & 3.56  & 6.45  & 4.53  & 6.18   & 6.34  & \textbf{7.45} & 6.50\\
        \hline
    \end{tabular}}
    \caption{Comparison of inference speed across different CPU (Unit: Tokens/Second) in an unlimited thread setting. $b(x)$ represents the bits per weight, where $x$ denotes specific value. "N/A" indicates that the tested CPU cannot host the specified model size with the given kernel. The token generation speed was determined by calculating the average results from 10 tests conducted across different devices using diverse methodologies.}
    \label{tab:detailed_com}
\end{table*}

\end{document}